\pdfoutput=1

\documentclass[11pt]{article}

\usepackage[]{acl}

\usepackage{times}
\usepackage{latexsym}
\usepackage{bm}

\usepackage[T1]{fontenc}

\usepackage[utf8]{inputenc}

\usepackage{microtype}

\usepackage{algorithm}
\usepackage{algorithmic}
\usepackage{url}            %
\usepackage{booktabs}       %
\usepackage{amsfonts}       %
\usepackage{nicefrac}       %
\usepackage{microtype}      %
\usepackage{xcolor}         %
\usepackage{lipsum}
\usepackage{arydshln}
\usepackage{pgfplots}
\usepackage[encapsulated]{CJK}
\usepackage{multirow}
\usepackage{mathrsfs}
\usepackage{mathtools}
\usepackage{subfigure}
\pgfplotsset{compat=1.15}
\usepackage[draft]{changes}

\title{Document-Level Relation Extraction with Sentences Importance Estimation and Focusing}

\author{Wang Xu\textsuperscript{\rm 1}, Kehai Chen\textsuperscript{\rm 1}, Lili Mou\textsuperscript{\rm 2}, Tiejun Zhao\textsuperscript{\rm 1} \\
  \textsuperscript{\rm 1}Harbin Institute of Technology, China \\
  \textsuperscript{\rm 2}Dept. Computing Science, Alberta Machine Intelligence Institute (Amii)\\
  University of Alberta, Canada \\
  xuwang@hit-mtlab.net, \{chenkehai,tjzhao\}@hit.edu.cn, doublepower.mou@gmail.com\\}

\begin{document}
\maketitle
\begin{abstract}
Document-level relation extraction~(DocRE) aims to determine the relation between two entities from a document of multiple sentences.
Recent studies typically represent the entire document by sequence- or graph-based models to predict the relations of all entity pairs. However, we find that such a model is not robust and exhibits bizarre behaviors: it predicts correctly when an entire test document is fed as input, but errs when non-evidence sentences are removed. To this end, we propose a Sentence Importance Estimation and Focusing (SIEF) framework for DocRE, where we design a sentence importance score and a sentence focusing loss, encouraging DocRE models to focus on evidence sentences. Experimental results on two domains show that our SIEF not only improves overall performance, but also makes DocRE models more robust. Moreover, SIEF is a general framework, shown to be effective when combined with a variety of base DocRE models.\footnote{The code is publicly available at \url{https://github.com/xwjim/SIEF}}
\end{abstract}

\section{Introduction}
\label{sec1}

Document-level relation extraction (DocRE) aims to predict entity relations across multiple sentences. It plays a crucial role in a variety of knowledge-based applications, such as question answering~\cite{sorokin-gurevych-2017-context} and large-scale knowledge graph construction~\cite{baldini-soares-etal-2019-matching}.
Different from sentence-level relation extraction~\cite{zeng-etal-2014-relation,xiao-liu-2016-semantic,Song_2019}, the supporting evidence in the DocRE setting may involve multiple sentences scattering in the document. 
Thus, DocRE is more a realistic setting, attracting increasing attention in the field of information extraction.

\begin{figure}[t!]
  \centering
  \includegraphics{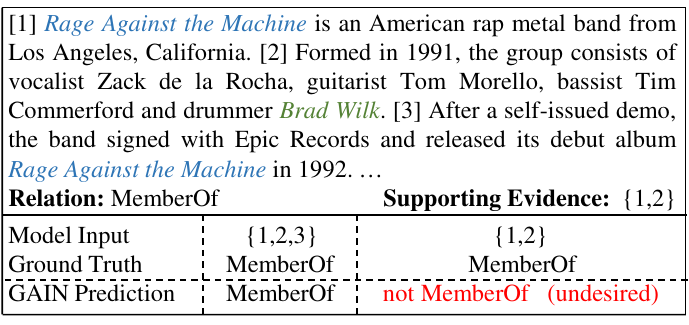}
  \caption{A DocRE model predicts correctly for an entire document, but errs when a non-evidence sentence is removed.}
  \label{fig1:example}
\end{figure}

Most recent DocRE studies use the entire document as a clue to predict the relations of all entity pairs without concerning where the evidence is located~\cite{Nan2020ReasoningWL,zeng-etal-2020-double,xu-etal-2021-discriminative,docred-rec}. 
However, one can identify the relation of a specific entity pair from a few sentences.
\citet{huang-etal-2021-three} show that irrelevant sentences in the document would hinder the performance of the model. 

Moreover, we observe that a DocRE model, trained on the entire document, may err when non-evidence sentences are removed.
In Figure~\ref{fig1:example}, for example, we need to identify the relation ``MemberOf'' between the entities \textit{Brad Wilk} and \textit{Rage Against the Machine}.
The evidence sentences are \{1,2\}, and humans can easily identify such a relation when reading sentences \{1,2\} only.
However, the recent DocRE model GAIN~\cite{zeng-etal-2020-double} identifies the relation ``MemberOf'' correctly from the entire document \{1,2,3\}, but predicts ``not MemberOf'' from sentences \{1,2\}. 
Intuitively, removing sentence \{3\} should not change the result, as this sentence does not provide information regarding whether ``MemberOf'' holds or not for the two entities.
Such model behaviors are undesired, because it shows that the model is not robust and lacks interpretability. 

To this end, we propose a novel \textbf{S}entence \text{I}mportance \textbf{E}stimation and \textbf{F}ocusing~(\textbf{SIEF}) framework to encourage the model to focus on evidence sentences for predicting the relation of an entity pair.
Specifically, we first evaluate the importance of each sentence by the difference between the output probabilities of the document with and without this sentence. If the predicted probability of a relation does not change, or even increases, when a sentence is removed, it typically indicates that the sentence is \textit{non-evidence}.
Then, we propose an auxiliary loss to encourage the model to produce the same output distribution, when the entire document is fed as input and when a non-evidence sentence is removed.
In this way, the model pays more attention to the evidence sentences for the classification.
Our SIEF method is a general framework that can be combined with different underlying DocRE models.

We evaluated the generality and effectiveness of our approach on the large-scale DocRED dataset~\cite{yao-etal-2019-docred}.
Experimental results show that the proposed approach combines well with various recent DocRE models and significantly improves the performance.
We further evaluated our approach on a dialogue relation extraction dataset, DialogRE~\cite{yu-etal-2020-dialogue}; our SIEF yields consistent improvement, showing the generality of our approach in different domains.

\section{Related Work}
\label{sec2}
Relation extraction (RE) can be categorized by its granularity, such as sentence-level~\cite{doddington-etal-2004-automatic,xu-etal-2016-improved,wei2019novel} and document-level~\cite{DBLP:journals/corr/abs-1810-05102,zhu-etal-2019-graph}.
Early work mainly focuses on sentence-level relation extraction. \newcite{pantel-pennacchiotti-2006-espresso} propose a rule-based approach, and \newcite{mintz-etal-2009-distant} manually design features for classifying relations. In the past several years, neural networks have become a prevailing approach for relation extraction~\cite{xu-etal-2015-classifying,Song_2019}.

Document-level relation extraction~(DocRE) is attracting increasing attention in the community, as it considers the interactions of entity mentions expressed in different sentences~\cite{Li2016cdr,yao-etal-2019-docred}. Compared with the sentence level, DocRE requires the model collecting and integrating inter-sentence information effectively. Recent efforts design sequence-based and graph-based models to address such a problem.

Sequence-based DocRE models encode a document by the sequence of words and/or sentences, for example, using the Transformer architecture~\cite{devlin-etal-2019-bert}.
\newcite{zhou2021atlop} argue that the Transformer attentions are able to extract useful contextual features across sentences for DocRE, and they adopt an adaptive threshold for each entity pair.
\newcite{ijcai2021-551} model DocRE as a semantic segmentation task and predict an entity-level relation matrix to capture local and global information.

Graph-based DocRE models abstract a document by graphical structures. For example, a node can be a sentence, a mention, and/or an entity; their co-occurrence is modeled by an edge. Then graph neural networks are applied to aggregate inter-sentence information~\cite{quirk-poon-2017-distant,Christopoulou2019ConnectingTD,zeng-etal-2020-double}. 
\newcite{zeng-etal-2020-double} construct double graphs, applying graph neural networks to mention--document graphs and performing path reasoning over entity graphs.
\newcite{xu-etal-2021-discriminative} explicitly incorporate logical reasoning, common-sense reasoning, and coreference reasoning into DocRE, based on both sequence and graph features.

Different from previous work, our paper proposes SIEF as a general framework that can be combined with various sequence-based and graph-based DocRE models. In our approach, we propose a sentence importance score and a sentence focusing loss to encourage the model to focus on evidence sentences, improving the robustness and the overall performance of DocRE models.

\begin{figure*}[ht!]
  \centering
  \includegraphics{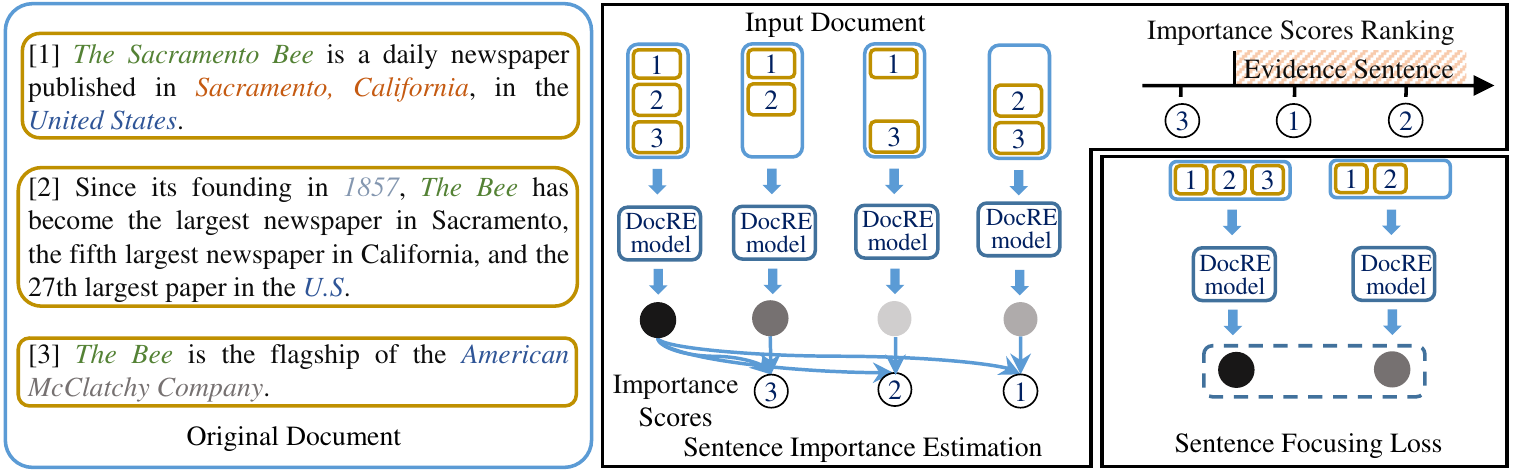}
  \caption{We estimate the sentence importance (for a specific entity pair and relation) by the difference of the classification probabilities with and without the sentence. Then, we encourage the DocRE model to predict the same probability when the entire document is fed as input and when a non-evidence sentence is removed.}
  \label{fig2:data_consistent}
\end{figure*}

\section{Problem Definition}
\label{sec3}
In this section, we present the formulation of document relation extraction (DocRE). Consider an unstructured document comprising $N$ sentences, $\mathbb{D}=\{\mathbf s_1, \mathbf s_2, \cdots, \mathbf  s_{N}\}$, where each sentence $\mathbf s_n$ is a sequence words. In a DocRE dataset, the document $\mathbb{D}$ is typically annotated with entity mentions, each mention (e.g., U.S.~and USA) labeled by its conceptual entity $e$ and its entity type (e.g., location).

A DocRE model $\bm F$ is usually formulated as multi-label classification~\cite{yao-etal-2019-docred}. $\bm F_j$ predicts  whether the $j$th relation holds for the $i$th marked entity pair in a document, given by
\begin{equation}
    P_{ij}=\bm F_j(\mathbb{D}, e_{i_h}, e_{i_t})= \Pr[r_{ij}=1| \mathbb{D}, e_{i_h}, e_{i_t}]
\label{eq1}
\end{equation}
where $e_{i_h}$ is the head entity and $e_{i_t}$ is the tail entity;  $r_{ij}\in\{0,1\}$ is the groundtruth label regarding entity pair $i$ and relation $j$.

To train the model, the binary cross-entropy loss is used as the objective for parameter estimation:
\begin{equation}
\begin{split}
\mathcal{L}_\text{rel}=-\sum_{\mathbb D \in \mathcal{C}}\sum_{i_h \neq i_t}\sum_{j \in \mathcal{R}} \{r_{ij} \log P_{ij}\hspace{1cm}\\\hspace{1cm}
+(1-r_{ij})\log(1-P_{ij})\}
\end{split}
\label{eq2}
\end{equation}
where $\mathcal{C}$ denotes the entire corpus and $\mathcal{R}$ denotes the set of relation types.

During inference, we obtain the relation(s) of a given entity pair by thresholding the predicted probabilities, following most previous work~\cite{yao-etal-2019-docred,zhou2021atlop}.

\section{Methodology}
\label{sec4}

In this section, we will describe our approach in detail.
The overview of our framework is shown in Figure~\ref{fig2:data_consistent}.
First, we describe the estimation of sentence importance in Section \ref{sec4-1}.
Sentences with low importance scores are treated as non-evidence.
Then, Sections~\ref{sec4-2} and \ref{sec4-3} present our approach that encourages the model to produce the same output distribution, when the entire document is fed as input and when  non-evidence sentences are removed.
 Section~\ref{sec4-4} further presents the architectures of DocRE models.

\subsection{Sentence Importance Estimation}
\label{sec4-1}
We estimate the importance of each sentence for a specific entity pair. Low-scored sentences will be treated as non-evidence, and in principle, can be removed without changing DocRE predictions. 

We propose a sentence importance score based on the DocRE predictions with and without the sentence in question. 
Our observation is that the relation extraction task is usually monotonic to evidence, i.e., (non-strictly) more relations will be predicted with more sentences. 
If we remove a sentence and the predicted probability of a relation decreases, then the sentence is likely to be the evidence.
If the predicted probability does not change, then the sentence is likely to be non-evidence. Moreover, the predicted probability may sometimes increase when a sentence is removed, in which case the DocRE model is not robust, as this violates monotonicity.

Formally, we consider removing one sentence at a time, and the document with the $n$th sentence removed is denoted by $\hat{\mathbb D}^{(-n)}=\{\mathbf s_1, \cdots, \mathbf s_{n-1}, \mathbf s_{n+1},\cdots, \mathbf s_N\}$.
For a DocRE model $\bm{F}$, we obtain the classification probabilities $P_{ij}=\bm F_j(\mathbb D,e_{i_h},e_{i_t})$ based on the original document, and $\hat{P}_{ij}^{(-n)}=\bm F_j(\hat{\mathbb D}^{(-n)},e_{i_h},e_{i_t})$ with sentence $n$ removed.

We propose the importance score as 
\begin{equation}
\begin{aligned}
g_{ij}^{(-n)}=P_{ij}\log\tfrac{P_{ij}}{\hat{P}_{ij}^{(-n)}}
\end{aligned}
\label{eq3}
\end{equation}

The formula appears similar to Kullback--Leibler (KL) divergence. 
However, we only take one term in the KL summation, because the KL divergence, albeit asymmetric in its two arguments, cannot model the increase or decrease of $\hat{P}_{ij}^{(-n)}$, whereas our $g^{(-n)}_{ij}$ is monotonically decreasing with $\hat{P}_{ij}^{(-n)}$. 
Compared with a naive difference or ratio between $P_{ij}$ and $\hat{P}^{(-n)}_{ij}$, we find that our KL-like score is more robust in the scale of ${P}_{ij}$ when determining non-evidence sentences.

We treat a sentence $n$ as \textit{non-evidence} if $g^{(-n)}_{ij}<\beta$ for a thresholding hyperparameter $\beta$.
The resulting set of non-evidence sentences is denoted by $\mathbb{K}_{ij}$ for the an entity pair $(e_{i_h},e_{i_t})$ and relation~$j$.

\subsection{Sentence Focusing Loss}
\label{sec4-2}
We propose a sentence focusing loss to encourage the model to produce the same output distribution when the entire document is fed as input and when non-evidence sentences are removed.

Ideally, the predicted probability should remain the same if we remove any combination of the sentences in $\mathbb{K}_{ij}$. Therefore, we penalize the extent to which the predicted probability is changed.

We propose the sentence focusing loss as:
\begin{equation}
\begin{aligned}
\mathcal{L}_{\text{sf}}=-\sum_{\mathbb D \in \mathcal{C}}\sum_{i_h \neq i_t}\sum_{j \in \mathcal{R}}\sum_{\mathbb J_{ij}\subseteq \mathbb K_{ij}}\{P_{ij}\log(\hat{P}^{(-\mathbb J_{ij})}_{ij})\\
+(1-P_{ij})\log(1-\hat{P}^{(-\mathbb J_i)}_{ij})\}
\end{aligned}
\label{eq4}
\end{equation}
where $\mathbb{J}_{ij}$ is a subset of $\mathbb{K}_{ij}$ and $\hat P_{ij}^{(-\mathbb J_{ij})} = \bm F_j(\mathbb D\backslash{\mathbb J_{ij}}, e_{i_h}, e_{i_t})$ is the predicted probability with $\mathbb J_{ij}$ removed from $\mathbb D$, and the total loss is $\mathcal{L}=(\mathcal{L}_{\textup{rel}}+\mathcal{L}_{\textup{sf}})/2$.

Essentially, our sentence focusing loss ensures $P_{ij}$ is close to $\hat P_{ij}^{(-\mathbb J_{ij})}$, which intuitively makes sense because non-evidence sentences should not affect the prediction. Our approach can also be thought of as a way of data augmentation.
However, compared with one-hot groundtruth labels, our sentence focusing loss works with soft labels $P_{ij}$ and $\hat P_{ij}^{(-\mathbb J_{ij})}$, which are believed to contain more information~\cite{kd44873}, and our gradient propagates to both $P_{ij}$ and $\hat P_{ij}^{(-\mathbb J_{ij})}$ for training. 

The calculation of Eqn.~\eqref{eq4} is time- and resource-consuming, because the number of the subsets $\mathbb J_{ij}$ grows combinatorially with the number of non-evidence sentences. 
Moreover it should be calculated repeatedly once the parameter of the model is updated.
To this end, we propose a simplified training strategy to approximate Eqn.~\eqref{eq4} in the next subsection.

\subsection{Training Strategy}
\label{sec4-3}
We propose a strategy to simplify the calculation and the training procedure.
Concretely, we only remove one non-evidence sentence in $\mathbb K_{ij}$ at a time instead of a subset of $\mathbb J_{ij}\subseteq \mathbb K_{ij}$, and we aggregate the effect of different non-evidence sentences by:
\begin{equation}
\begin{aligned}
\mathcal{L}_{\text{sf}}=-\sum_{\mathbb D \in \mathcal{C}}\sum_{n=1}^N\sum_{i_h \neq i_t}\sum_{j \in \mathcal{R}}\mathbb{I}(g^{(-n)}_{ij}<\beta) \\
\{P_{ij}\log(\hat{P}^{(-n)}_{ij})+(1-P_{ij})\log(1-\hat{P}^{(-n)}_{ij})\}
\end{aligned}
\label{eq6}
\end{equation}
where $\mathbb{I}$ is the indicator function. Essentially, we linearly approximate the combination of multiple non-evidence sentences in \eqref{eq4}  by an outer summation. In this way, the number of terms does not grow combinatorially, but linearly w.r.t.~$N$.

In implementation, we further simply the summation over $n$ by Monte Carlo sampling of a randomly selected sentence~$n$ in each gradient update.
The loss is reformulated as follows:
\begin{equation}
\begin{aligned}
\mathcal{L}_{\text{sf}}=-\sum_{\mathbb D \in \mathcal{C}}\sum_{i_h \neq i_t}\sum_{j \in \mathcal{R}}\mathbb{I}(g^{(-n)}_{ij}<\beta) \\
\{P_{ij}\log(\hat{P}^{(-n)}_{ij})+(1-P_{ij})\log(1-\hat{P}^{(-n)}_{ij})\}
\end{aligned}
\label{eq7}
\end{equation}

As seen, we need to forward the base models twice in each update, with and without the sentence $n$. 
\citet{huang-etal-2021-three} propose a similar idea but train different entity pairs in a document based on different sets of sentences; all sentence are processed repeatedly among entity pairs in a document. Their approach is much slower than ours.

To sum up, the proposed SIEF framework identifies non-evidence sentences and penalizes the difference of predicted probabilities when a non-evidence sentence is removed. 
Our approach is a generic framework and can be adapted to various DocRE model easily, without introducing extra parameters into the model.

\subsection{DocRE Model Architectures}
\label{sec4-4}
Our SIEF can be applied to various base DocRE models.
To evaluate its generality, we consider the following recent models. 

\textbf{BiLSTM}~\cite{yao-etal-2019-docred}\footnote{https://github.com/thunlp/DocRED\label{rep_docred}}. A bi-directional long short term memory (BiLSTM) encodes the document, and an entity is representated by BiLSTM's hidden states, averaged over entity mentions. The head and tail entity representations are fed to a multi-layer perceptron (MLP) for relation extraction.

\textbf{BERT$_\text{base}$}~\cite{devlin-etal-2019-bert}\footnote{https://github.com/DreamInvoker/GAIN\label{rep_gain}}. A pre-trained language model is used for document encoding.

\textbf{HeterGSAN}~\cite{docred-rec}\footnote{https://github.com/xwjim/DocRE-Rec\label{rep_rec}}. HeterGSAN is a recent graph-based DocRED model, which constructs a heterogeneous graph of sentence, mention, and entity nodes; it uses graph neural networks for relation extraction.

\textbf{GAIN}~\cite{zeng-etal-2020-double}\textsuperscript{\ref {rep_gain}}. GAIN constructs two graphs: mention--document graphs and entity graphs, and performs graph and path reasoning over the two graphs separately. When combining our SIEF with GAIN, we achieve the best performance among all the base models with SIEF on DocRED. Thus, we will explain this model in more detail.

\begin{figure}[!t]
  \centering
  \includegraphics{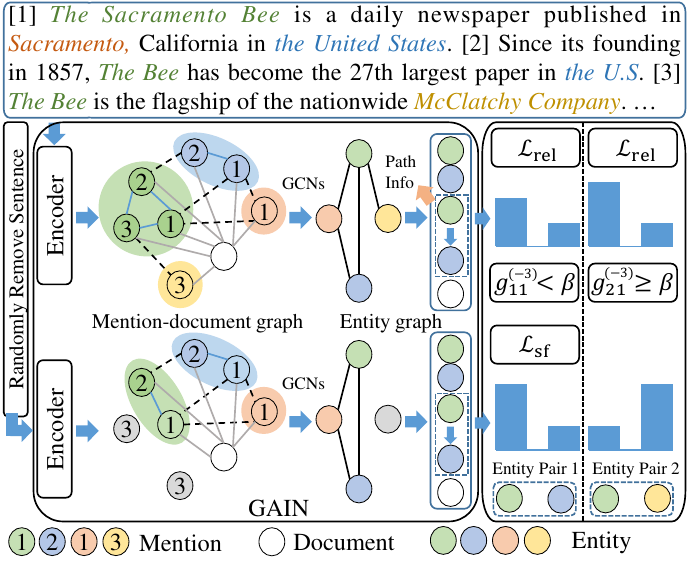}\vspace{-.2cm}
  \caption{The model architecture of GAIN with SIEF. A sentence is randomly removed from the document. The corresponding nodes and edges are removed from the mention--document graph and the entity graph.}\vspace{-.2cm}
  \label{fig3:model_arch}
\end{figure}

Essentially, a node in the mention--document graph is either a mention or a document. The mentions are connected to its document, and two mentions are connected if they co-occur in one sentence. In the entity graph, two entities are connected if they are mentioned in one sentence. To classify the relation, GNN is applied to the mention--document graph, enhanced with path information in the entity graph, shown in Figure~\ref{fig3:model_arch}.

When combining SIEF with GAIN, we randomly remove one sentence from the document. The corresponding nodes and edges are removed in the GAIN's graphs.
Then we obtain the output probabilities with and without the sentence, $P_{ij}$ and $\hat{P}^{(-n)}_{ij}$, separately.
If the sentence important score $g_{ij}^{(-n)}$ in Eqn.~\eqref{eq3} is below a threshold $\beta$, the sentence is treated as non-evidence for the entity pair $(e_{i_h},e_{i_t})$ and relation $j$. We apply the sentence focusing loss Eqn.~\eqref{eq4} to improve the robustness.

For prediction, we apply the trained DocRE model to the entire document, because with our approach the model is already robust when non-evidence sentences are presented. Empirical results will show that our SIEF consistently improves the performance of base DocRE models. 

\section{Experiments}
\label{sec5}
\subsection{Setup}
\label{sec5-1}
\noindent \textbf{Datasets.} DocRED is a large-scale human-annotated dataset for document-level relation extraction~\cite{yao-etal-2019-docred}.
The dataset is constructed from Wikipedia and Wikidata, containing 3053 documents for training, 1000 for development, and 1000 for test. In total, it has 132,375 entities and 56,354 relational facts in 96 relation types. 
More than 40\% of the relational facts require reasoning over multiple sentences.
The standard evaluation metrics are F1 and Ign~F1 \cite{yao-etal-2019-docred,zeng-etal-2020-double}, where Ign~F1 refers to the F1 score excluding the relational facts in the training set.

We also evaluated our approach on \mbox{DialogRE} \cite[V2,][]{yu-etal-2020-dialogue}, which contains 36 relation types, 17 of which are interpersonal. We followed the standard split with 1073 training dialogues, 358 validation, and 357 test. Following \newcite{yu-etal-2020-dialogue}, we report macro F1 scores in both the standard and conversational settings; the latter is denoted by F1$_c$.

\textbf{Competing Methods.} We experimented our SIEF on a number of base models, namely, BiLSTM, BERT$_\text{base}$, HeterGSAN, and GAIN (Section~\ref{sec4-4}). 
These base models are all considered for comparison. 

For DocRED, we consider additional competing methods: 
\textbf{Two Phase}~\cite{Wang2019FinetuneBF}, which first predicts whether the entity pair has a relation and then predicts the relation type;
\textbf{LSR}~\cite{Nan2020ReasoningWL}, which constructs the graph by inducing a latent document-level graph;
\textbf{Reconstructor}~\cite{docred-rec}, which encourages the model to reconstruct a reasoning path during training;
\textbf{DRN}~\cite{xu-etal-2021-discriminative}, which considers different reasoning skills explicitly and uses graph representation and context representation to model the reasoning skills;
\textbf{ATLOP}~\cite{zhou2021atlop}, which aggregates contextual information by the Transformer attentions and adopts an adaptive threshold for different entity pairs;
and \textbf{DocuNet}~\cite{ijcai2021-551}, which models DocRE as a semantic segmentation task.

For DialogRE, we followed \newcite{yu-etal-2020-dialogue} and considered \textbf{BERT} and \textbf{BERT}$_\text{s}$ for comparison,\footnote{https://github.com/nlpdata/dialogre\label{rep_dialog}}
where \textbf{BERT}$_\text{s}$ prevents a model from overfitting by replacing of the interpersonal augment with a special token.

\textbf{Implementation Details}. We use the repositories\textsuperscript{\ref {rep_docred},\ref{rep_gain},\ref{rep_rec},\ref{rep_dialog}} of base models to implement our approach.
We mostly followed the standard hyperparameters used in the base models. 
Our SIEF has one hyperparameter $\beta$ in Eqn.~\eqref{eq6}. It was set to 0.8, and Section~\ref{sec5-2} presents the effect of tuning $\beta$.

\subsection{Results and Analyses}
\label{sec5-2}
\begin{table}[!t]
\begin{center}
\scalebox{.69}{
\begin{tabular}{l|l|l|l|l}
\hline \hline
\multicolumn{1}{c|}{\multirow{2}{*}{Model}} & \multicolumn{2}{c|}{Dev} & \multicolumn{2}{c}{Test} \\ \cline{2-5} 
                       \multicolumn{1}{c|}{}                       & Ign F1        & F1       & Ign F1        & F1        \\ \hline
\multicolumn{5}{c}{\textit{DocRE Systems with GloVe}}                                    \\ \hline
 LSR~\cite{Nan2020ReasoningWL}          & 48.82 & 55.17 & 52.15 & 54.18 \\ 
 Reconstructor~\cite{docred-rec}        & 54.25 & 55.70 & 53.25 & 55.13   \\  
 DRN~\cite{xu-etal-2021-discriminative} & 54.61 & 56.49 & 54.35 & 56.33     \\\hline
  BiLSTM~\cite{yao-etal-2019-docred}            & 48.87 & 50.94 & 48.78 & 51.06 \\
 \;\;\;+\textbf{SIEF}                 & 52.08 & 54.20 & 51.03 & 53.22 \\   \cdashline{1-5}
 HeterGSAN~\cite{docred-rec}            & 52.17 & 54.40 & 52.07 & 53.52\\
 \;\;\;+\textbf{SIEF}                 & 54.49 & 56.30 & 53.94 & 55.85\\  \cdashline{1-5}
 GAIN~\cite{zeng-etal-2020-double}      & 53.05 & 55.29 & 52.66 & 55.08 \\  
 \;\;\;+\textbf{SIEF}                 & 55.07 & 56.96 & 54.72 & 56.75\\  \hline \hline 
\multicolumn{5}{c}{\textit{DocRE Systems with BERT$_\text{base}$}}                                              \\ \cline{1-5} 
Two-Phase~\cite{Wang2019FinetuneBF}     & -     & 54.42 & -     & 53.92\\
LSR~\cite{Nan2020ReasoningWL}           & 52.43 & 59.00 & 56.97 & 59.05 \\ 
Reconstructor~\cite{docred-rec}     & 58.13 & 60.18 & 57.12 & 59.45 \\ 
DRN~\cite{xu-etal-2021-discriminative} & 59.33 & 61.39 & 59.15 & 61.37 \\
ATLOP~\cite{zhou2021atlop}     & 59.22 & 61.09 & 59.31 & 61.30 \\
DocuNet~\cite{ijcai2021-551} & 59.86 & 61.83 & {\textbf{59.93}} & 61.86  \\ \hline
BERT$_\text{base}$~\cite{ye-etal-2020-coreferential}          & 54.63 & 56.77 & 53.93 & 56.27 \\ 
\;\;\;+\textbf{SIEF}                 & 57.13 & 59.11 & 57.87 & 58.93\\  \cdashline{1-5}
HeterGSAN~\cite{docred-rec}         & 57.00 & 59.13 & 56.21 & 58.54 \\
\;\;\;+\textbf{SIEF}             & 57.99 & 60.04 & 57.93 & 60.02   \\  \cdashline{1-5}
GAIN~\cite{zeng-etal-2020-double}   & 59.14 & 61.22 & 59.00 & 61.24 \\ 
\;\;\;+\textbf{SIEF}             & \textbf{59.82} & \textbf{62.24} & {59.87} & \textbf{62.29}   \\ \hline \hline
\end{tabular}}
\end{center}
\caption{Results on the development and test sets of the DocRE dataset. 
Bold indicates the best performance.}
\label{tab:mainresult} 
\end{table}

\begin{table}[!t]
\centering
\resizebox{0.9\linewidth}{!}{
\begin{tabular}{l|cc|cc}
\hline
\multirow{2}{*}{Model} & \multicolumn{2}{c|}{Dev} & \multicolumn{2}{c}{Test} \\ \cline{2-5} 
                       & F1     & F1$_c$  & F1   & F1$_c$       \\ \hline
BERT~\cite{yu-etal-2020-dialogue}                   & 60.6   & 55.4    & 58.5 & 53.2          \\
\;\;\;+\textbf{SIEF}             & 61.4   & 57.6    & 59.9 & 56.1         \\ \hline
BERT$_\text{s}$~\cite{yu-etal-2020-dialogue}               & 63.0   & 57.3    & 61.2 & 55.4             \\
\;\;\;+\textbf{SIEF}             & \textbf{64.3}   & \textbf{60.6}  & \textbf{61.8} & \textbf{58.4}          \\ \hline
\end{tabular}}
\caption{Results on DialogRE.}
\label{tab:dialogre}
\end{table}

\textbf{Main results.}
Table~\ref{tab:mainresult} presents the detailed results on the development and test sets of the DocRED dataset. 
We first compare DocRE systems with GloVe embeddings~\cite{yao-etal-2019-docred}. We see that the proposed SIEF method significantly improves the performance of all base models, including the sequence model (i.e., BiLSTM) and graph models (i.e., HeterGSAN and GAIN); the average improvement is 2.05 points in terms of test F1. This shows that SIEF is compatible with both sequence and graph models, indicating the generality and effectiveness of the proposed method.

For the DocRE system with BERT$_\text{base}$, SIEF also consistently improves the base models, showing that SIEF is complementary to the modern BERT architecture. Especially, combining SIEF and GAIN~\cite{zeng-etal-2020-double} with BERT$_\text{base}$ encoding yields state-of-the-art performance in terms of F1.

We further conducted experiments on the DialogRE dataset, and compare our approach with the BERT baselines in \newcite{yu-etal-2020-dialogue}. As seen, the results are consistent with the improvement on DocRED, as our SIEF largely improves F1 and F1$_c$ for both base models. This further confirms the generality of our approach in different domains.

In the rest of this section, we present in-depth analyses to better understand our model with DocRED as the testbed. All base models use GloVe embeddings as opposed to BERT due to efficiency concerns.
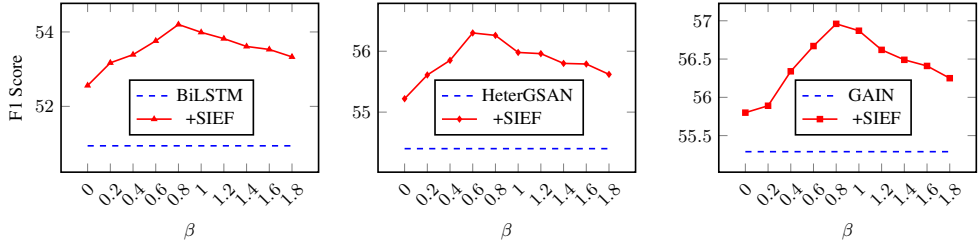
\begin{figure*}[ht!]\centering
\resizebox{.8\linewidth}{!}{
    \begin{minipage}[b]{0.32\linewidth}
    \centering
		\pgfplotsset{height=4.5cm,width=6.0cm,compat=1.14,every axis/.append style={thick},every axis legend/.append style={at={(1,0.13)}},legend columns=1}
		\begin{tikzpicture}
		\tikzset{every node}=[font=\small]
		\begin{axis}
		[enlargelimits=0.13, tick align=inside, 
		xtick={0,0.2,0.4,0.6,0.8,1.0,1.2,1.4,1.6,1.8},
		x tick label style={rotate=45},
		ymin=50.7,
		ymax=54.3,
		legend style={at={(0.75,0.55)}},
		ylabel={F1 Score},xlabel={$\beta$},font=\small]
		\addplot [dashed,mark size=1.2pt,mark options={solid,mark color=blue}, color=blue] coordinates
		{(0,50.94)(0.2,50.94)(0.4,50.94)(0.6,50.94)(0.8,50.94)(1.0,50.94)(1.2,50.94)(1.4,50.94)(1.6,50.94)(1.8,50.94)};
		\addlegendentry{\small BiLSTM}
		\addplot [sharp plot,mark=triangle*,mark size=1.2pt,mark options={solid,mark color=red}, color=red] coordinates
		{(0,52.56)(0.2,53.17)(0.4,53.39)(0.6,53.76)(0.8,54.20)(1.0,53.99)(1.2,53.82)(1.4,53.61)(1.6,53.53)(1.8,53.33)};
		\addlegendentry{\small +SIEF}
		\end{axis}
		\end{tikzpicture}
		    \end{minipage}
    \hspace{0.5cm}
        \begin{minipage}[b]{0.32\linewidth}
    \centering
    \pgfplotsset{height=4.5cm,width=6.0cm,compat=1.14,every axis/.append style={thick},every axis legend/.append style={at={(1,0.13)}},legend columns=1}
		\begin{tikzpicture}
		\tikzset{every node}=[font=\small]
		\begin{axis}
		[enlargelimits=0.13, tick align=inside, 
		xtick={0,0.2,0.4,0.6,0.8,1.0,1.2,1.4,1.6,1.8},
		x tick label style={rotate=45},
		ymin=54.3,
		ymax=56.5,
		legend style={at={(0.80,0.55)}},
		xlabel={$\beta$},font=\small]
		\addplot [dashed,mark size=1.2pt,mark options={solid,mark color=blue}, color=blue] coordinates
		{(0,54.40)(0.2,54.40)(0.4,54.40)(0.6,54.40)(0.8,54.40)(1.0,54.40)(1.2,54.40)(1.4,54.40)(1.6,54.40)(1.8,54.40)};
		\addlegendentry{\small HeterGSAN}
		\addplot [sharp plot,mark=diamond*,mark size=1.2pt,mark options={solid,mark color=red}, color=red] coordinates
		{(0,55.22)(0.2,55.61)(0.4,55.85)(0.6,56.30)(0.8,56.26)(1.0,55.98)(1.2,55.96)(1.4,55.80)(1.6,55.79)(1.8,55.62)};
		\addlegendentry{\small +SIEF\;\;\;\;}
		\end{axis}
		\end{tikzpicture}
		\end{minipage}
    \hspace{0.35cm}
    \begin{minipage}[b]{0.32\linewidth}
    \centering
    \pgfplotsset{height=4.5cm,width=6.0cm,compat=1.14,every axis/.append style={thick},every axis legend/.append style={at={(1,0.13)}},legend columns=1}
		\begin{tikzpicture}
		\tikzset{every node}=[font=\small]
		\begin{axis}
		[enlargelimits=0.13, tick align=inside, 
		xtick={0,0.2,0.4,0.6,0.8,1.0,1.2,1.4,1.6,1.8},
		x tick label style={rotate=45},
		ymin=55.25,
		ymax=57.0,
		legend style={at={(0.75,0.55)}},
		xlabel={$\beta$},font=\small]
		\addplot [dashed,mark size=1.2pt,mark options={solid,mark color=blue}, color=blue] coordinates
		{(0,55.29)(0.2,55.29)(0.4,55.29)(0.6,55.29)(0.8,55.29)(1.0,55.29)(1.2,55.29)(1.4,55.29)(1.6,55.29)(1.8,55.29)};
		\addlegendentry{\small GAIN}
		\addplot [sharp plot,mark=square*,mark size=1.2pt,mark options={solid,mark color=red}, color=red] coordinates
		{(0,55.80)(0.2,55.89)(0.4,56.34)(0.6,56.67)(0.8,56.96)(1.0,56.87)(1.2,56.62)(1.4,56.49)(1.6,56.41)(1.8,56.25)};
		\addlegendentry{\small \;\;+SIEF}
		\end{axis}
		\end{tikzpicture}
		\end{minipage}}\vspace{-.35cm}
		\caption{\label{fig:hypertheta}Performances of the classification (in F1 scores) on the development set of different hyperparameter $\beta$ in Eqn.~\eqref{eq7} during the training.}\vspace{-.2cm}
\end{figure*}

\textbf{Intra- and Inter-Sentence Performance.}
We breakdown the relation classification performance into intra-sentence reasoning and inter-sentence reasoning. Ideally, if only one sentence is needed to determine the relation of an entity pair, then it belongs to the intra-sentence category; if two or more sentences are needed, then it belongs to the inter-sentence category. We follow~\newcite{Nan2020ReasoningWL} and approximate it by checking whether two entities are mentioned in one sentence.

The results are shown in Table~\ref{tab:reasonperformance}. SIEF again consistently improves base models in terms of both Intra-F1 and Inter-F1. However, the improvement on Intra-F1 is larger than that on Inter-F1. This is because our SIEF encourages the model to focus on evidence by removing one sentence at a time, but does not explicitly model sentence relations. Based on this analysis, we plan to extend the SIEF framework with multi-sentence DocRE reasoning in our future work.

\begin{table}[!t]
\centering
\resizebox{0.8\linewidth}{!}{
\begin{tabular}{l|ll}
\hline
Model             & Intra-F1     & Inter-F1    \\ \hline
BiLSTM            & 57.05 & 43.49   \\
\;\;\;\;+\textbf{SIEF}      & 60.56 ${}_{(\Delta=+3.51)}$ & 45.96${}_{(\Delta=+2.47)}$   \\ \hline
HeterGSAN         & 61.79 & 47.06   \\
\;\;\;\;+\textbf{SIEF}      & 63.01 ${}_{(\Delta=+1.22)}$ & 48.11${}_{(\Delta=+1.05)}$   \\ \hline
GAIN              & 61.67 & 48.77   \\
\;\;\;\;\textbf{+SIEF}      & \textbf{63.21} ${}_{(\Delta=+1.54)}$ & \textbf{48.98}${}_{(\Delta=+0.21)}$   \\ \hline
\end{tabular}}
\caption{Results of Intra-F1 results and Infer-F1 on development set of DocRED. The difference is compared between SIEF and the respective base model.}
\label{tab:reasonperformance}
\end{table}

\textbf{Performance of predicting evidence sentences.} 
In our paper, we propose a sentence importance score to measure how much a sentence contributes to the classification without using additional annotation. We evaluate such performance in Table~\ref{tab:performanceevidence} by Precision, Recall, and F1 scores against manually annotated evidence sentences that are provided in the dataset. In this analysis, we do not perform relation prediction, but concern about entity pairs knowingly having certain relations.
Specifically, for entity pair $(e_{i_h},e_{i_t})$ with relation $j$, we calculate the importance score $g_{ij}^{(-n)}$ for each sentence and cut off evidence/non-evidence sentences with a threshold based on the development F1 score. 

As seen, all base models achieve above 60\% F1, suggesting that the proposed importance score is indeed indicative for predicting evidence and non-evidence sentences.

With the proposed SIEF framework, the performance improves for all metrics, with an average improvement of 2.95 F1 points across three base models.
This further verifies that our SIEF framework not only improves relation extraction performance, but also is able to better detect evidence and non-evidence sentences, which is important for the interpretability of machine learning models.

\begin{table}[!t]
\centering
\resizebox{0.8\linewidth}{!}{
\begin{tabular}{l|lll}
\hline
Model     & Precision   & Recall & F1    \\ \hline
BiLSTM            & 60.14 & 68.41  & 64.01 \\
\;\;\;\;+\textbf{SIEF}      & 65.00 & 67.99  & 66.46${}_{(\Delta=+2.45)}$ \\ \hline
HeterGSAN         & 65.40 & 70.95  & 68.06 \\
\;\;\;\;+\textbf{SIEF}      & 71.40 & 70.21  & 70.80${}_{(\Delta=+2.74)}$ \\ \hline
GAIN              & 65.28 & 71.17  & 68.10 \\
\;\;\;\;+\textbf{SIEF}      & \textbf{71.94} & \textbf{71.60}  & \textbf{71.77}${}_{(\Delta=+3.67)}$ \\ \hline
\end{tabular}}\vspace{-.1cm}
\caption{Results of the evidence prediction on the development set of DocRED.}\vspace{-.2cm}
\label{tab:performanceevidence}
\end{table}

\begin{figure}[!t]\vspace{-.3cm}
  \centering
  \includegraphics{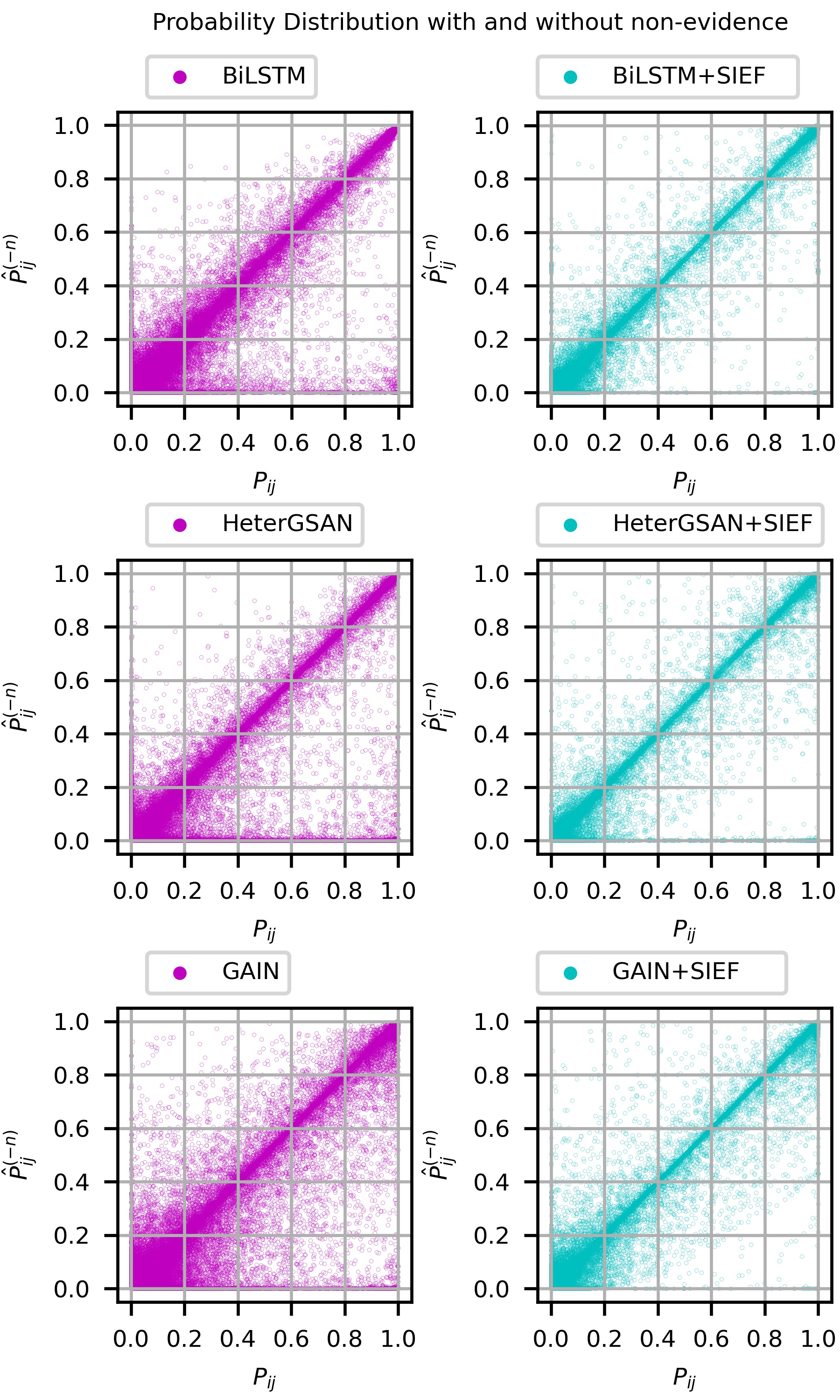} \vspace{-.6cm}
  \caption{Robustness of DocRE models.}\vspace{-.2cm}
  \label{fig:data_point_dis}
\end{figure}

\textbf{Robustness of DocRE models.}
We further investigate the robustness of DocRE models by showing the difference between the predicted distributions with and without non-evidence sentences. We show  in Figure~\ref{fig:data_point_dis}  the scatter plots of the probability $P$ based on the entire document and the probability $\hat{P}_{ij}^{(-n)}$ with a random non-evidence sentence removed. 

As shown in the figure, the points of the base models (left magenta plots) scatters over a wider range, whereas our SIEF training (right cyan plots) makes them more concentrated on the diagonal, indicating that the prediction $P_{ij}$ on the entire document is mostly the same as $\hat{P}_{ij}^{(-n)}$ with a non-evidence removed. This shows the robustness of SIEF-trained models, as they are less sensitive to non-evidences sentences for DocRE.

\textbf{Analysis on hyperparameter $\beta$.} 
Our SIEF framework has one hyperaparameter $\beta$ that controls how strict we treat a sentence as evidence or non-evidence (Section~\ref{sec4-3}). We analyze the effect of~$\beta$ in Figure~\ref{fig:hypertheta}. 

As seen, our SIEF approach consistently benefits the base models with a large range of $\beta$ values. Intuitively, if $\beta$ is too small, very few sentences will be treated as non-evidence and our sentence focusing loss is less effective; if $\beta$ is too large, it has a high false positive rate of non-evidence sentences. Empirically, a moderate $\beta$ around (0.6--0.8) yields the highest performance. From the plots, we also see that our hyperparameter $\beta$ is insensitive to the base models, justifying our design of Eqn.~\eqref{eq3}.

\textbf{Sentence importance score VS other heuristics.}
To investigate the effectiveness of our sentence importance score in~Eqn.~\eqref{eq3}, we compare it with several alternative heuristics: 1) We randomly select half of the sentences as the non-evidence set, denoted by \textbf{Rand}; and 2) We consider the non-evidence set as the sentences without entity mentions, denoted by \textbf{NoMention}.

The results of the performance in terms of F1 and Ign F1 on the development set are shown in Table~\ref{tab:sentence_evalution}.
As seen, the simple heuristic Rand outperforms the base model, as Rand can be thought of as noisy data augmentation. The NoMention heuristic outperforms Rand, as sentences without entity mentions are more likely to be non-evidence. Moroever, SIEF is superior to both Rand and NoMention, showing that our sentence importance scores is a more effective indicator of evidence and non-evidence sentences.

\begin{table}[!t]
\centering
\scalebox{0.75}{
\begin{tabular}{l|cc|cc|cc}
\hline
\multicolumn{1}{c|}{\multirow{2}{*}{Method}} & \multicolumn{2}{c|}{BiLSTM}                           & \multicolumn{2}{c|}{HeterGSAN}                        & \multicolumn{2}{c}{GAIN}                             \\ \cline{2-7} 
\multicolumn{1}{c|}{}      & \multicolumn{1}{c}{Ign F1} & \multicolumn{1}{c|}{F1} & \multicolumn{1}{c}{Ign F1} & \multicolumn{1}{c|}{F1} & \multicolumn{1}{c}{Ign F1} & \multicolumn{1}{c}{F1} \\ \hline
Base           & 48.87 & 50.94 & 52.17 & 54.40 & 53.05 & 55.29 \\
+\textbf{SIEF}      & \textbf{52.08} & \textbf{54.20} & \textbf{54.49} & \textbf{56.30} & \textbf{55.07} & \textbf{56.96} \\ \cdashline{1-7}
+Rand      & 50.63 & 52.63 & 52.75 & 54.70 & 53.41 & 55.63  \\
+NoMention\!\!       & 51.56 & 53.79 & 54.07 & 55.95 & 54.66 & 56.52  \\ \hline
\end{tabular}}
\caption{Results of our approach and other heuristics.}
\label{tab:sentence_evalution}
\end{table}

\textbf{Our sentence focusing loss VS learning from groundtruth.}
\begin{table}[!t]
\centering
\scalebox{0.75}{
\begin{tabular}{l|cc|cc|cc}
\hline
\multicolumn{1}{c|}{\multirow{2}{*}{Method}} & \multicolumn{2}{c|}{BiLSTM}                           & \multicolumn{2}{c|}{HeterGSAN}                        & \multicolumn{2}{c}{GAIN}                             \\ \cline{2-7} 
\multicolumn{1}{c|}{}      & \multicolumn{1}{c}{Ign F1} & \multicolumn{1}{c|}{F1} & \multicolumn{1}{c}{Ign F1} & \multicolumn{1}{c|}{F1} & \multicolumn{1}{c}{Ign F1} & \multicolumn{1}{c}{F1} \\ \hline
Base           & 48.87 & 50.94 & 52.17 & 54.40 & 53.05 & 55.29 \\
+\textbf{SIEF}      & \textbf{52.08} & \textbf{54.20} & \textbf{54.49} & \textbf{56.30} & \textbf{55.07} & \textbf{56.96}  \\  \cdashline{1-7}
+GTruth\!\!      & 50.36 & 52.56 & 52.65 & 54.69 & 53.75 & 55.87 \\ \hline
\end{tabular}}
\caption{Comparing our sentence focusing loss with learning from groundtruth labels (denoted by GTruth).}
\label{tab:hard_label}
\end{table}
We encourage the DocRE models to generate consistent output probabilities with and without non-evidence (Section~\ref{sec4-2}) by a cross-entropy loss between two soft distributions $P_{ij}$ and $\hat{P}_{ij}^{(-n)}$.
To investigate the effect of such a sentence focusing loss, we compare it with an alternative choice: we learn $\hat{P}_{ij}^{(-n)}$ directly from the groundtruth label $r_{ij}$.

Table~\ref{tab:hard_label} shows the results on the development set in terms of F1 and Ign F1.
As seen, both methods can improve the performance of the base models. This confirms that removing non-evidence sentences can serve as a way of data augmentation, boosting the performance of DocRE models.
Moreover, we observe that our sentence focusing loss is better than learning from the groundtruth labels, showing that the soft predictions provide more information than one-hot labels, consistent with knowledge distillation literature~\cite{kd44873}. 

\begin{figure}[!t]
  \centering
  \includegraphics{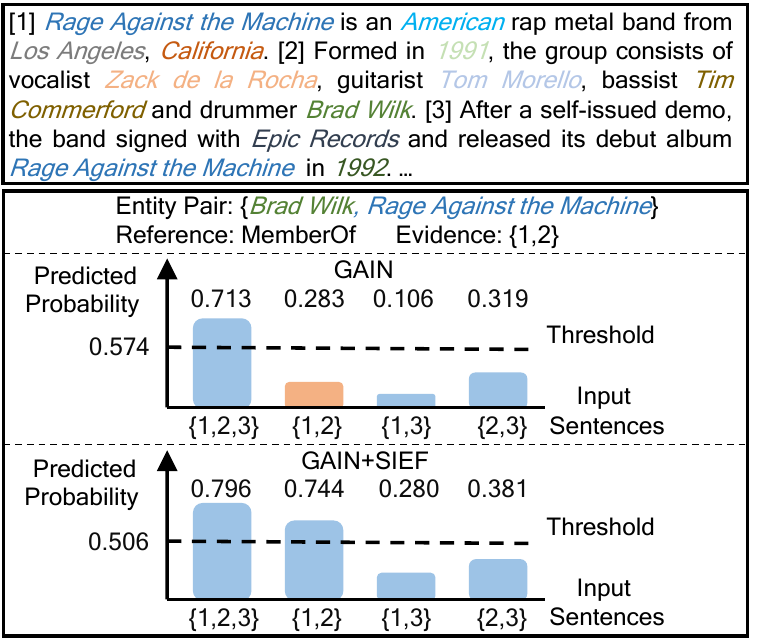}
  \caption{Case Study.}
  \label{fig:casestudy}
\end{figure}

\textbf{Case Study.}
Figure~\ref{fig:casestudy} shows a case study of GAIN and GAIN+SIEF models. 
For the entity pair (\textit{Brad Wilk}, \textit{Rage Against the Machine}), both GAIN and GAIN+SIEF predicts the relation ``MemberOf'', which is consistent with the reference.
We see that Sentence 3 is non-evidence, and in principle, it should not affect DocRE prediction in this case. However, the base GAIN model makes a wrong prediction ``not MemberOf'', as the predicted probability is below the threshold, which is determined by validation based on predicted binary probabilities of all relations. By contrast, our SIEF model is able to make correct predictions when different non-evidence sentences are removed, demonstrating its robustness.

\section{Conclusion}
\label{sec6}
In this paper, we propose a novel Sentence Information Estimation and Focusing (SIEF) approach to document relation extraction (DocRE). We design a sentence importance score and a sentence focusing loss to encourage the model to focus on evidence sentences. The proposed SIEF is a general framework, and can be combined with various base DocRE models. 
Experimental results show that SIEF consistently improves the performance of base models in different domains, and that it improves the robustness of DocRE models.

\section*{Acknowledgments}
We are grateful to the anonymous reviewers and meta reviewers for their insightful comments and suggestions.
This work is funded by the National Key Research and Development Program of China (No.~2020AAA0108000). Lili Mou is supported in part by the Natural Sciences and Engineering Research Council of Canada (NSERC) under grant
No.~RGPIN2020-04465, the Amii Fellow Program, the Canada CIFAR AI Chair Program, a UAHJIC project, a donation from DeepMind, and Compute Canada (www.computecanada.ca).

\bibliography{acl_latex}

\begin{thebibliography}{29}
\expandafter\ifx\csname natexlab\endcsname\relax\def\natexlab#1{#1}\fi

\bibitem[{Baldini~Soares et~al.(2019)Baldini~Soares, FitzGerald, Ling, and
  Kwiatkowski}]{baldini-soares-etal-2019-matching}
Livio Baldini~Soares, Nicholas FitzGerald, Jeffrey Ling, and Tom Kwiatkowski.
  2019.
\newblock \href {https://doi.org/10.18653/v1/P19-1279} {Matching the blanks:
  Distributional similarity for relation learning}.
\newblock In \emph{Proceedings of the Annual Meeting of the Association for
  Computational Linguistics}, pages 2895--2905.

\bibitem[{Christopoulou et~al.(2019)Christopoulou, Miwa, and
  Ananiadou}]{Christopoulou2019ConnectingTD}
Fenia Christopoulou, Makoto Miwa, and Sophia Ananiadou. 2019.
\newblock \href {https://doi.org/10.18653/v1/D19-1498} {Connecting the dots:
  Document-level neural relation extraction with edge-oriented graphs}.
\newblock In \emph{Proceedings of the Conference on Empirical Methods in
  Natural Language Processing and the International Joint Conference on Natural
  Language Processing}, pages 4925--4936.

\bibitem[{Devlin et~al.(2019)Devlin, Chang, Lee, and
  Toutanova}]{devlin-etal-2019-bert}
Jacob Devlin, Ming-Wei Chang, Kenton Lee, and Kristina Toutanova. 2019.
\newblock \href {https://doi.org/10.18653/v1/N19-1423} {{BERT}: Pre-training of
  deep bidirectional transformers for language understanding}.
\newblock In \emph{Proceedings of the Conference of the North {A}merican
  Chapter of the Association for Computational Linguistics}, pages 4171--4186.

\bibitem[{Doddington et~al.(2004)Doddington, Mitchell, Przybocki, Ramshaw,
  Strassel, and Weischedel}]{doddington-etal-2004-automatic}
George Doddington, Alexis Mitchell, Mark Przybocki, Lance Ramshaw, Stephanie
  Strassel, and Ralph Weischedel. 2004.
\newblock \href {http://www.lrec-conf.org/proceedings/lrec2004/pdf/5.pdf} {The
  automatic content extraction ({ACE}) program {--} tasks, data, and
  evaluation}.
\newblock In \emph{Proceedings of the Fourth International Conference on
  Language Resources and Evaluation}, pages 837--840.

\bibitem[{Gupta et~al.(2019)Gupta, Rajaram, Sch{\"{u}}tze, and
  Runkler}]{DBLP:journals/corr/abs-1810-05102}
Pankaj Gupta, Subburam Rajaram, Hinrich Sch{\"{u}}tze, and Thomas~A. Runkler.
  2019.
\newblock \href {https://doi.org/10.1609/aaai.v33i01.33016513} {Neural relation
  extraction within and across sentence boundaries}.
\newblock In \emph{Proceedings of the Association for the Advance of Artificial
  Intelligence}, pages 6513--6520.

\bibitem[{Hinton et~al.(2015)Hinton, Vinyals, and Dean}]{kd44873}
Geoffrey Hinton, Oriol Vinyals, and Jeffrey Dean. 2015.
\newblock \href {http://arxiv.org/abs/1503.02531} {Distilling the knowledge in
  a neural network}.
\newblock In \emph{Proceedings of the Annual Conference on Neural Information
  Processing Systems Deep Learning and Representation Learning Workshop}.

\bibitem[{Huang et~al.(2021)Huang, Zhu, Feng, Ye, Lai, and
  Zhao}]{huang-etal-2021-three}
Quzhe Huang, Shengqi Zhu, Yansong Feng, Yuan Ye, Yuxuan Lai, and Dongyan Zhao.
  2021.
\newblock \href {https://doi.org/10.18653/v1/2021.acl-short.126} {Three
  sentences are all you need: Local path enhanced document relation
  extraction}.
\newblock In \emph{Proceedings of the Annual Meeting of the Association for
  Computational Linguistics and the International Joint Conference on Natural
  Language Processing}, pages 998--1004.

\bibitem[{Li et~al.(2016)Li, Sun, Johnson, Sciaky, Wei, Leaman, Davis,
  Mattingly, Wiegers, and Lu}]{Li2016cdr}
Jiao Li, Yueping Sun, Robin~J. Johnson, Daniela Sciaky, Chih-Hsuan Wei, Robert
  Leaman, Allan~Peter Davis, Carolyn~J. Mattingly, Thomas~C. Wiegers, and
  Zhiyong Lu. 2016.
\newblock \href {https://doi.org/https://doi.org/10.1093/database/baw068}
  {{BioCreative V CDR} task corpus: A resource for chemical disease relation
  extraction}.
\newblock \emph{Database: The Journal of Biological Databases and Curation}.

\bibitem[{Mintz et~al.(2009)Mintz, Bills, Snow, and
  Jurafsky}]{mintz-etal-2009-distant}
Mike Mintz, Steven Bills, Rion Snow, and Daniel Jurafsky. 2009.
\newblock \href {https://aclanthology.org/P09-1113} {Distant supervision for
  relation extraction without labeled data}.
\newblock In \emph{Proceedings of the Joint Conference of the Annual Meeting of
  the Association for Computational Linguistics and the International Joint
  Conference on Natural Language Processing}, pages 1003--1011.

\bibitem[{Nan et~al.(2020)Nan, Guo, Sekulic, and Lu}]{Nan2020ReasoningWL}
Guoshun Nan, Zhijiang Guo, Ivan Sekulic, and Wei Lu. 2020.
\newblock \href {https://doi.org/10.18653/v1/2020.acl-main.141} {Reasoning with
  latent structure refinement for document-level relation extraction}.
\newblock In \emph{Proceedings of the Annual Meeting of the Association for
  Computational Linguistics}, pages 1546--1557.

\bibitem[{Pantel and Pennacchiotti(2006)}]{pantel-pennacchiotti-2006-espresso}
Patrick Pantel and Marco Pennacchiotti. 2006.
\newblock \href {https://doi.org/10.3115/1220175.1220190} {{E}spresso:
  Leveraging generic patterns for automatically harvesting semantic relations}.
\newblock In \emph{Proceedings of the International Conference on Computational
  Linguistics and Annual Meeting of the Association for Computational
  Linguistics}, pages 113--120.

\bibitem[{Quirk and Poon(2017)}]{quirk-poon-2017-distant}
Chris Quirk and Hoifung Poon. 2017.
\newblock \href {https://www.aclweb.org/anthology/E17-1110} {Distant
  supervision for relation extraction beyond the sentence boundary}.
\newblock In \emph{Proceedings of the Conference of the European Chapter of the
  Association for Computational Linguistics}, pages 1171--1182.

\bibitem[{Song et~al.(2019)Song, Zhang, Gildea, Yu, Wang, and Su}]{Song_2019}
Linfeng Song, Yue Zhang, Daniel Gildea, Mo~Yu, Zhiguo Wang, and Jinsong Su.
  2019.
\newblock \href {https://doi.org/10.18653/v1/d19-1020} {Leveraging dependency
  forest for neural medical relation extraction}.
\newblock \emph{Proceedings of the Conference on Empirical Methods in Natural
  Language Processing and the International Joint Conference on Natural
  Language Processing}.

\bibitem[{Sorokin and Gurevych(2017)}]{sorokin-gurevych-2017-context}
Daniil Sorokin and Iryna Gurevych. 2017.
\newblock \href {https://doi.org/10.18653/v1/D17-1188} {Context-aware
  representations for knowledge base relation extraction}.
\newblock In \emph{Proceedings of the Conference on Empirical Methods in
  Natural Language Processing}, pages 1784--1789.

\bibitem[{Wang et~al.(2019)Wang, Focke, Sylvester, Mishra, and
  Wang}]{Wang2019FinetuneBF}
Hong Wang, Christfried Focke, Rob Sylvester, Nilesh Mishra, and William W.~J.
  Wang. 2019.
\newblock \href {https://arxiv.org/abs/1909.11898} {Fine-tune {BERT} for
  {DocRED} with two-step process}.
\newblock \emph{arXiv preprint arXiv:1909.11898}.

\bibitem[{Wei et~al.(2020)Wei, Su, Wang, Tian, and Chang}]{wei2019novel}
Zhepei Wei, Jianlin Su, Yue Wang, Yuan Tian, and Yi~Chang. 2020.
\newblock \href {https://www.aclweb.org/anthology/2020.acl-main.136} {A novel
  cascade binary tagging framework for relational triple extraction}.
\newblock In \emph{Proceedings of the Annual Meeting of the Association for
  Computational Linguistics}, pages 1476--1488.

\bibitem[{Xiao and Liu(2016)}]{xiao-liu-2016-semantic}
Minguang Xiao and Cong Liu. 2016.
\newblock \href {https://aclanthology.org/C16-1119} {Semantic relation
  classification via hierarchical recurrent neural network with attention}.
\newblock In \emph{Proceedings of the International Conference on Computational
  Linguistics: Technical Papers}, pages 1254--1263.

\bibitem[{Xu et~al.(2021{\natexlab{a}})Xu, Chen, and
  Zhao}]{xu-etal-2021-discriminative}
Wang Xu, Kehai Chen, and Tiejun Zhao. 2021{\natexlab{a}}.
\newblock \href {https://doi.org/10.18653/v1/2021.findings-acl.144}
  {Discriminative reasoning for document-level relation extraction}.
\newblock In \emph{Findings of the Annual Meeting of the Association for
  Computational Linguistics}, pages 1653--1663.

\bibitem[{Xu et~al.(2021{\natexlab{b}})Xu, Chen, and Zhao}]{docred-rec}
Wang Xu, Kehai Chen, and Tiejun Zhao. 2021{\natexlab{b}}.
\newblock \href {https://ojs.aaai.org/index.php/AAAI/article/view/17667}
  {Document-level relation extraction with reconstruction}.
\newblock In \emph{Proceedings of the Association for the Advance of Artificial
  Intelligence}, 16, pages 14167--14175.

\bibitem[{Xu et~al.(2016)Xu, Jia, Mou, Li, Chen, Lu, and
  Jin}]{xu-etal-2016-improved}
Yan Xu, Ran Jia, Lili Mou, Ge~Li, Yunchuan Chen, Yangyang Lu, and Zhi Jin.
  2016.
\newblock \href {https://aclanthology.org/C16-1138} {Improved relation
  classification by deep recurrent neural networks with data augmentation}.
\newblock In \emph{Proceedings of the International Conference on Computational
  Linguistics: Technical Papers}, pages 1461--1470.

\bibitem[{Xu et~al.(2015)Xu, Mou, Li, Chen, Peng, and
  Jin}]{xu-etal-2015-classifying}
Yan Xu, Lili Mou, Ge~Li, Yunchuan Chen, Hao Peng, and Zhi Jin. 2015.
\newblock \href {https://doi.org/10.18653/v1/D15-1206} {Classifying relations
  via long short term memory networks along shortest dependency paths}.
\newblock In \emph{Proceedings of the Conference on Empirical Methods in
  Natural Language Processing}, pages 1785--1794.

\bibitem[{Yao et~al.(2019)Yao, Ye, Li, Han, Lin, Liu, Liu, Huang, Zhou, and
  Sun}]{yao-etal-2019-docred}
Yuan Yao, Deming Ye, Peng Li, Xu~Han, Yankai Lin, Zhenghao Liu, Zhiyuan Liu,
  Lixin Huang, Jie Zhou, and Maosong Sun. 2019.
\newblock \href {https://doi.org/10.18653/v1/P19-1074} {{D}oc{RED}: A
  large-scale document-level relation extraction dataset}.
\newblock In \emph{Proceedings of the Annual Meeting of the Association for
  Computational Linguistics}, pages 764--777.

\bibitem[{Ye et~al.(2020)Ye, Lin, Du, Liu, Li, Sun, and
  Liu}]{ye-etal-2020-coreferential}
Deming Ye, Yankai Lin, Jiaju Du, Zhenghao Liu, Peng Li, Maosong Sun, and
  Zhiyuan Liu. 2020.
\newblock \href {https://doi.org/10.18653/v1/2020.emnlp-main.582}
  {Coreferential reasoning learning for language representation}.
\newblock In \emph{Proceedings of the Conference on Empirical Methods in
  Natural Language Processing}, pages 7170--7186.

\bibitem[{Yu et~al.(2020)Yu, Sun, Cardie, and Yu}]{yu-etal-2020-dialogue}
Dian Yu, Kai Sun, Claire Cardie, and Dong Yu. 2020.
\newblock \href {https://doi.org/10.18653/v1/2020.acl-main.444} {Dialogue-based
  relation extraction}.
\newblock In \emph{Proceedings of the Annual Meeting of the Association for
  Computational Linguistics}, pages 4927--4940.

\bibitem[{Zeng et~al.(2014)Zeng, Liu, Lai, Zhou, and
  Zhao}]{zeng-etal-2014-relation}
Daojian Zeng, Kang Liu, Siwei Lai, Guangyou Zhou, and Jun Zhao. 2014.
\newblock \href {https://www.aclweb.org/anthology/C14-1220} {Relation
  classification via convolutional deep neural network}.
\newblock In \emph{Proceedings of the International Conference on Computational
  Linguistics: Technical Papers}, pages 2335--2344.

\bibitem[{Zeng et~al.(2020)Zeng, Xu, Chang, and Li}]{zeng-etal-2020-double}
Shuang Zeng, Runxin Xu, Baobao Chang, and Lei Li. 2020.
\newblock \href {https://doi.org/10.18653/v1/2020.emnlp-main.127} {Double graph
  based reasoning for document-level relation extraction}.
\newblock In \emph{Proceedings of the Conference on Empirical Methods in
  Natural Language Processing}, pages 1630--1640.

\bibitem[{Zhang et~al.(2021)Zhang, Chen, Xie, Deng, Tan, Chen, Huang, Si, and
  Chen}]{ijcai2021-551}
Ningyu Zhang, Xiang Chen, Xin Xie, Shumin Deng, Chuanqi Tan, Mosha Chen, Fei
  Huang, Luo Si, and Huajun Chen. 2021.
\newblock \href {https://doi.org/10.24963/ijcai.2021/551} {Document-level
  relation extraction as semantic segmentation}.
\newblock In \emph{Proceedings of the Thirtieth International Joint Conference
  on Artificial Intelligence}, pages 3999--4006.

\bibitem[{Zhou et~al.(2021)Zhou, Huang, Ma, and Huang}]{zhou2021atlop}
Wenxuan Zhou, Kevin Huang, Tengyu Ma, and Jing Huang. 2021.
\newblock \href {https://ojs.aaai.org/index.php/AAAI/article/view/17717}
  {Document-level relation extraction with adaptive thresholding and localized
  context pooling}.
\newblock In \emph{Proceedings of the Association for the Advance of Artificial
  Intelligence}, 16, pages 14612--14620.

\bibitem[{Zhu et~al.(2019)Zhu, Lin, Liu, Fu, Chua, and
  Sun}]{zhu-etal-2019-graph}
Hao Zhu, Yankai Lin, Zhiyuan Liu, Jie Fu, Tat-Seng Chua, and Maosong Sun. 2019.
\newblock \href {https://doi.org/10.18653/v1/P19-1128} {Graph neural networks
  with generated parameters for relation extraction}.
\newblock In \emph{Proceedings of the Annual Meeting of the Association for
  Computational Linguistics}, pages 1331--1339.

\end{thebibliography}
\bibliographystyle{acl_natbib}

\appendix
\clearpage

\end{document}